\documentclass[runningheads]{llncs}
\usepackage{graphicx}
\usepackage{comment}
\usepackage{lipsum}
\usepackage{listings}
\usepackage{float}
\usepackage{booktabs}
\usepackage{cite}
\usepackage{adjustbox}
\usepackage[symbol]{footmisc}

\renewcommand{\thefootnote}{\fnsymbol{footnote}}

\newcommand\blfootnote[1]{%
  \begingroup
  \renewcommand\thefootnote{}\footnote{#1}%
  \addtocounter{footnote}{-1}%
  \endgroup
}

\begin{document}
\title{A Search Engine for Scientific Publications: a Cybersecurity Case Study}
\titlerunning{A Search Engine for Scientific Publications}
\author{Nuno Oliveira\orcidID{0000-0002-5030-7751} \and
Norberto Sousa\orcidID{0000-0003-2919-4817} \and
Isabel Praça\orcidID{0000-0002-2519-9859}}
\authorrunning{N. Oliveira and N. Sousa et al.}
\institute{Research Group on Intelligent Engineering and Computing for Advanced Innovation and Development (GECAD), Porto School of Engineering (ISEP), 4200-072 Porto, Portugal \\ 
\email{\{nunal,norbe,icp\}@isep.ipp.pt}}

\maketitle              % typeset the header of the contribution
\begin{abstract}
Cybersecurity is a very challenging topic of research nowadays, as digitalization increases the interaction of people, software and services on the Internet by means of technology devices and networks connected to it. The field is broad and has a lot of unexplored ground under numerous disciplines such as management, psychology, and data science. Its large disciplinary spectrum and many significant research topics generate a considerable amount of information, making it hard for us to find what we are looking for when researching a particular subject. This work proposes a new search engine for scientific publications which combines both information retrieval and reading comprehension algorithms to extract answers from a collection of domain-specific documents. The proposed solution although being applied to the context of cybersecurity exhibited great generalization capabilities and can be easily adapted to perform under other distinct knowledge domains.

\keywords{Natural Language Processing \and Deep Learning \and Cybersecurity \and Question Answering System \and Reading Comprehension}
\end{abstract}

\section{Introduction}

\blfootnote{\textbf{Acknowledgements}: The present work has been developed under the EUREKA ITEA3 Project CyberFactory\#1 (ITEA-17032) and Project CyberFactory\#1PT (ANI|P2020 40124) co-funded by Portugal 2020.} Cybersecurity is a neoteric field that emerged out of the latest advances in computer science \cite{review1}. Although there is not yet a consensual agreement between the scientific community across the whole scope of cybersecurity research topics, some works have tried to systematize research categories \cite{review1,review2}, being one of them related to data science applications. 

The recent developments in software, hardware, and network topologies contributed to more complex systems such as Cyber-Physical Systems (CPS) in which the capabilities of computing,  communications, and data storage are used to monitor physical and cyber entities \cite{cps}. Furthermore, these advances can also be translated into more sophisticated cyberattacks comprised of multiple attack vectors. Hence, the complex nature of cyber threats and the need to progressively adapt security systems to the most relevant ones makes the application of Artificial Intelligence (AI) a promising technology to use for increased cybersecurity \cite{wirkuttis2017artificial}.

Being cybersecurity such a hot research topic nowadays, with so many different applications, it is hard to efficiently find answers to specific topics in the wide amount of existing scientific publications. Natural Language Processing (NLP) methods, namely Reading Comprehension (RC) algorithms, can give a substantial contribution to solving the introduced problem. However, the ability to read a text and then answer questions about it is a very difficult task for machines \cite{squad}.

Over the last few years, the introduction of reliable data collections such as the Stanford question answering dataset (SQuAD) \cite{squad} and the development of deep learning methods based on transformer architectures \cite{vaswani2017attention} such as BERT \cite{devlin2019bert} and RoBERTa \cite{liu2019roberta} have contributed to major improvements in the field of RC. Nevertheless, it is not feasible to apply these algorithms directly to huge amounts of text due to computational limits and performance issues. To overcome this problem, Information Retrieval (IR) methods \cite{bassil2012survey} can be used to measure the relevance of a given document to a given question providing a filter to find only relevant data and narrowing down the search space.

This work proposes a novel Question Answering (Q\&A) system for the cybersecurity research context in which one can place a domain-related question and expect a direct answer retrieved from a set of scientific publications. This system uses a combination of IR and RC methods to perform the task described above and provides the results in a user-friendly web interface.  Although the fact that the cybersecurity context is considered for the case study, the proposed method can be easily applied to many other different domains.

This work is organized in multiple sections that can be detailed as follows. Section \ref{section:related-work} provides an overview of current information retrieval and reading comprehension algorithms and applications. Section \ref{section:proposed-solution} describes the proposed solution, detailing both the software architecture and employed algorithms. In Section \ref{section:case-study}, our solution is applied to the case study and the obtained results are presented and discussed. Section \ref{section:conclusion} provides a summary of the main conclusions that can be drawn from this research and appoints further research topics to be addressed in the future.

\section{Related Work}
\label{section:related-work}
Due to the multiple domains intelligent QA systems are connected to, we will analyze the literature on multiple different subjects.
One of such subjects is text mining and document ranking systems, of which the internet and search engines are a great example \cite{singh2009comparative}. 
Taking into account the scope of our work, we investigated weighing methods such as Term Frequency - Inverse Document Frequency (TF-IDF), Dense Passage Retrievers (DPR), and word embeddings.

In \cite{qaiser2018text}, Shahzad Qaiser \textit{et al.}, employs a TF-IDF ranking system to several web pages in order to compare results. TF-IDF is the most utilized weighting scheme for web searches of information retrieval and text mining \cite{Beel2016}. The author also points TF-IDF's biggest issue, which is not identifying different tenses of words. In the same manner, Joel L. Neto \textit{et al.} in \cite{neto2000document} employs a modified version of TF-IDF, TF-ISF, applying stemming to reduce the impact of this classification method's weaknesses.

In \cite{karpukhin2020dense}, Karpukhin and Oğus \textit{et al.}, utilized the standard BERT pre-trained model and a DPR in a dual encoder architecture achieving state of the art results. Their DPR exceeds BM25's capabilities by far, namely a more than 20\% increase in top-5 accuracy (65.2\%). Their results for end-to-end QA accuracy also improved on ORQA, the first open-retrieval question answering system, introduced in \cite{lee2019latent} by Lee \textit{et al.}, in the natural questions dataset \cite{kwiatkowski2019natural}.

Regarding word embedding, in which a document's words are mapped as vectors in a continuous vector space, words with similar meanings will be closer to one another, aiding in dimensionality reduction \cite{8258123}. In \cite{mikolov2013distributed}, Tomas Mikolov \textit{et al.} demonstrates the application of a skip-gram model, a more computational efficient architecture, to mapping words to a vectorial space, and the same model but focusing on phrases. 

On the other hand, the Q\&A task involves the search for relationships and meaning between entities. Due to the nature of language, this search becomes extremely complex, given that context can change the meaning of any sequence of words. In NLP, the key to solve entity-related tasks is to create a model to learn the optimal way of entity representation. 

Ordinarily, each entity in the Knowledge Base (KB) is assigned an embedding vector, capturing information in it. Due to the scope restriction of this method, entities that are outside of the KB are not represented and therefore any model built on top of it performs poorly.

To solve this issue, Contextual Word Representations (CWR) are employed with generalized word representations that serve multiple purposes. These CWRs are based on the transformer architecture, most notably BERT \cite{devlin2019bert} and following improvements such as RoBERTa \cite{liu2019roberta} that perform extremely well in a wide range of NLP tasks such as document classification and entanglement, sentiment analysis, question answering, sentence similarity, etc. These representations are obtained by training a model on a large-scale corpus (ex: Wikipedia) and can then be transferred to other network-based models, allowing them to improve search-related tasks such as the relevant question context as shown by Wei Yang \textit{et al.} in \cite{yang2019end} where, in an end-to-end QA system, the integration of BERT outperformed previous implementations by significant margins. %\cite{fu2020bert}.

% @article{fu2020bert,
%   title={BERT for Question Answering on BioASQ},
%   author={Fu, Eric R and Djoko, Rikel and Mansor, Maysam and Slater, Robert},
%   journal={SMU Data Science Review},
%   volume={3},
%   number={3},
%   pages={3},
%   year={2020}
% }

\section{Proposed Solution}
\label{section:proposed-solution}

To solve the introduced problem we built a prototype using the python programming language on top of the haystack framework \cite{haystack}. The system was designed as a client-server architecture with two main components, the front-end, a web-based graphical interface that can be accessed by the users and the back-end, a RESTful API that exposes the use cases of our solution through several endpoints. Additionally, there is also an SQLite database which is used to store preprocessed scientific articles.

The back-end side of our application can also be further detailed into two distinct modules, a web-crawler, which is integrated with arXiv.org API so that it can fetch scientific articles in real time, and a search engine, which combines two distinct NLP methods, a retriever and a reader, to build a pipeline that is able to find candidate answers in our corpus to user-specified questions. The described architecture is represented in Figure 1.

\begin{figure}[H]
\centering
\includegraphics[width=8.5 cm]{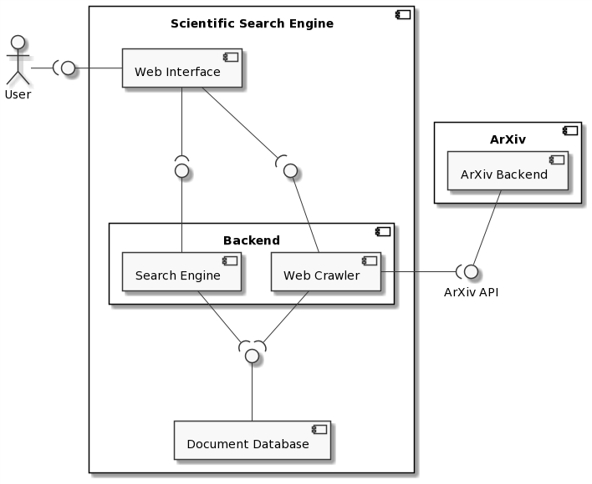}
\caption{Solution Architecture.}
\end{figure}

The proposed system regards three main use cases that can be described as follows:

\begin{itemize}
    \item \textbf{UC1 - Download Scientific Articles}: The user specifies a given search topic and the maximum number of articles to be downloaded. The crawler tries to find articles related to the specified topic and downloads all articles until the maximum limit is reached. Then, the documents are preprocessed - empty lines are removed, consecutive whitespaces are truncated and pdf headers and footers are discarded. After preprocessing, the text of each document is split into several search chunks of 500 words (respecting sentence boundary) so that the search process can be optimal. Finally, each resulting chunk is indexed, along with the document meta-data, in the document database.
    \item \textbf{UC2 - Consult Database Summary}: In the main dashboard of the graphical interface a summary of the document database content is displayed so that the user can keep track of the changes to the available corpus. The database summary is comprised of several data points such as the number of downloaded articles, search chunks and document categories.
    \item \textbf{UC3 - Find Candidate Answers}: The user places a question to the system and specifies several search parameters such as a category filter, the number of candidate answers to be displayed, $c$, and the maximum number of relevant search chunks to be found by the retriever, $k$. The system will first execute the retriever, a TF-IDF-based retriever which will return the most relevant $k$ chunks. Then, the reader, a RoBERTa model, will try to find the best $c$ answers in the selected $k$ chunks according to a confidence metric.
\end{itemize}

The described solution is quite generic since it is easy to enrich the search corpus with the contents of scientific publications of different subjects due to the execution of \textbf{UC1}. However, for this concrete implementation it would only be possible to find articles stored in the arXiv.org repository. Nonetheless, this feature can be easily expanded by integrating the existing web crawler with other scientific repositories. 

Regarding \textbf{UC3}, the proposed NLP pipeline is also quite broad, the retriever, TF-IDF is not context-specific and can easily be used for multiple domains. On the other hand, the reader, RoBERTa, requires training examples comprising different questions and answers. To overcome this limitation, we opted to use a model that was pre-trained on the SQuAD dataset \cite{roberta-deepset}. This data collection comprises over 100,000 examples of questions posed by crowdworkers on a set of Wikipedia articles \cite{squad} resembling a good benchmark dataset for training and evaluating general-purpose extractive Q\&A machine learning models. The RoBERTa model employed in our solution, \cite{roberta-deepset}, achieved an exact match score of approximately 79.97\% and an f1-score of 83.00\% under this testbed. In our experiments, the search engine performed quite competently being able to find interesting answers to several questions that were placed regarding the cybersecurity domain.

It is possible to further improve the proposed solution by adding new functionalities regarding the database management, namely, to perform listings of downloaded articles accordingly to a combination of search criteria, to manually import a given scientific article and to delete unwanted articles.

\subsection{Pipeline Description}

All steps of the search pipeline are described in this section.

\subsubsection{Retriever}\label{retriever}
In order to search through relevant information, a TF-IDF retriever was put in place. It is a numerical statistic that is intended to reflect how important a given word is to a document in a corpus. 
\begin{equation}
    TFIDF_{i,d} = tf_{i,d} \cdot idf_{i}
\end{equation}
In the scientific question and answering domain, it is expected that the queries will have lexical overlap with their answers, making this algorithm a good searcher of relevant information.

\subsubsection{Reader}
\begin{figure}[H]
\centering
\includegraphics[width=7 cm]{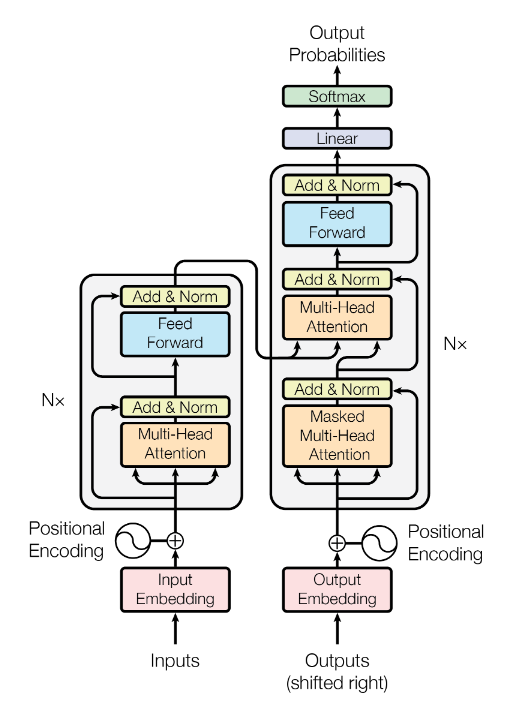}
\caption{The Transformer Architecture \cite{vaswani2017attention}.}
\label{fig:transformer}
\end{figure}

Another critical step of our pipeline is the question understanding step. Here we need to be able to properly understand the question at hand. By being able to properly model it in such a way that it can then be passed through the pipeline and improving the chances of getting not only accurate but also relevant answers.\
For this step, we use a FARM reader coupled with the RoBERTa \cite{liu2019roberta} language model which works alongside the retriever and parses the candidate documents provided.\
RoBERTa is an iteration of the BERT \cite{devlin2019bert} language model whose architecture is based on the Transformer architecture, Figure \ref{fig:transformer}. This new architecture disregards recurrence and convolutions from the usual encoder-decoder models and instead focuses on several types of attention mechanisms. It introduces several novelties such as sclaled-dot product attention, multi-head attention and positional encoding. At each time step the output of the decoder stack is fed back to the decoder similarly to how the outputs of previous time steps are used as hidden states in Recurrent Neural Networks (RNN) \cite{vaswani2017attention}.

RoBERTa is also trained on a much larger corpus than BERT and as a result, achieves significant performance gains. \

\section{Case Study}
\label{section:case-study}

Despite the usefulness and generalization of our solution, which allows it to be applied to numerous topics, for our case study we have decided to focus on a current and challenging research topic - cybersecurity.
For this reason we compiled a list of keywords related to that topic that we used to find relevant articles to build our search corpus. For each keyword we obtained a number of articles as shown in Table \ref{tab:corpus}. After removing the corrupted/unparsable documents and duplicates, our corpus totalled 821 articles.
	 
\begin{table}[H]
\caption{Corpus composition.}
\centering
\footnotesize
\begin{tabular}{lr}
\midrule
Adversarial Attack          & 200   \\
Attack Detection            & 175   \\
Cyberphysical Systems       & 200   \\
Cybersecurity               & 129   \\
Intrusion Detection Systems & 130   \\
                                    \\
Total Used                  & 834   \\
Corrupted                   & -6    \\
Duplicates                  & -7    \\
                                    \\
Total Articles in Corpus    & 821   \\
\bottomrule
\end{tabular}
\label{tab:corpus}
\end{table}
	 
Each one of these articles was downloaded and processed as per the pipeline indicated in the previous section. After processing, the articles were split into chunks of 500 words while taking into account sentence continuity.
With the finalization of this step, our corpus was composed of 12827 search chunks from 821 different articles of about 36 categories.

\subsection{Results}

\begin{figure}[H]
\centering
\includegraphics[width=10 cm]{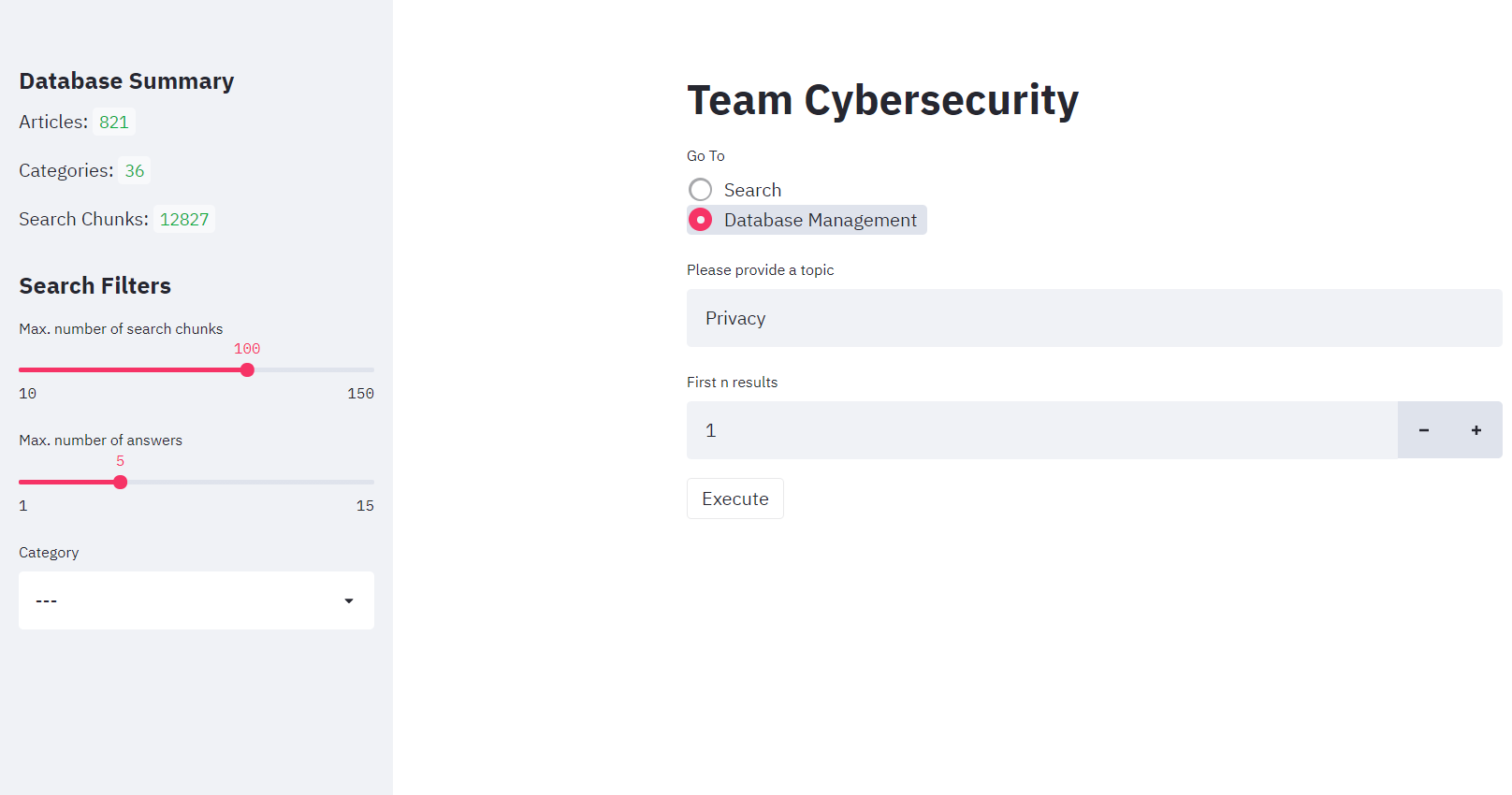}
\caption{Dashboard.}
\label{fig:dash}
\end{figure}

The introduced solution has a main dashboard, on the left some search configuration sliders and database related information is located. In the middle there are two buttons to navigate between the database management and search engine functionalities. The described interface is presented in Figure \ref{fig:dash}.

\begin{figure}[H]
\centering
\includegraphics[width=10 cm]{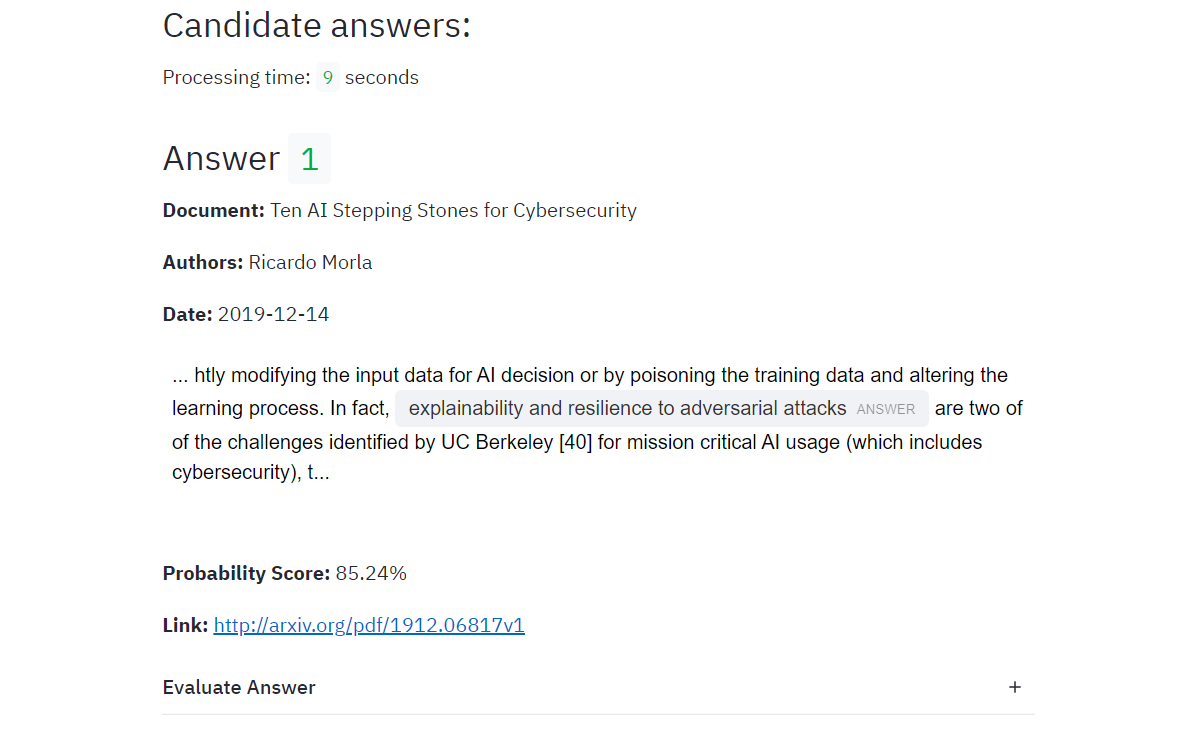}
\caption{Question Example.}
\label{fig:qa}
\end{figure}

With the corpus prepared, it is then possible to start asking questions. By asking: "What are the challenges of AI?", the most interesting candidate answer is presented in Figure \ref{fig:qa}, due to its high probability (confidence) score. This answer is highlighted in its surrounding context, accompanied by additional information such as title, authors, publishing date, and a link to the article itself.

As the question is vague in nature, and the prepared corpus is geared more towards cybersecurity instead of AI, the obtained answer "explainability and resilience to adversarial attacks" also tends to the cybersecurity side of AI, due to the nature of the used article \cite{Morla2019TenAS}.

Another example is the question,
"What are the main challenges of cybersecurity research?" which yielded interesting results. The first answer correctly quotes \cite{2021arXiv210103564K} and responds with "lack of adequate evaluation/test environments that utilize up-to-date datasets, variety of testbeds while adapting unified evaluation methods", while the second answer builds on the first one with "lack of research methodology standards" \cite{Gardner2019UsingCC}.

Finally, by asking "Which machine learning models are commonly used?" we obtain "Naïve Bayes, SVM, KNN, and decision trees" from \cite{cmc.2021.013852} and virtually the same answer "Support Vector Machine, Decision Trees, Fuzzy Logic, BayesNet and Naïve Bayes" from \cite{SHAH2018157}.

The quality of the responses found is directly connected to the contents of the corpus. This can be remedied by populating the corpus with more articles pertaining to a given topic or adding a new topic entirely.
For this we can access the database management functionality, and specify a given search topic and the maximum number of documents to be downloaded. These will be directly fetched from arXiv.org, preprocessed and indexed alongside their metadata in the document database.

\begin{figure}[H]
\centering
\includegraphics[width=10.5 cm]{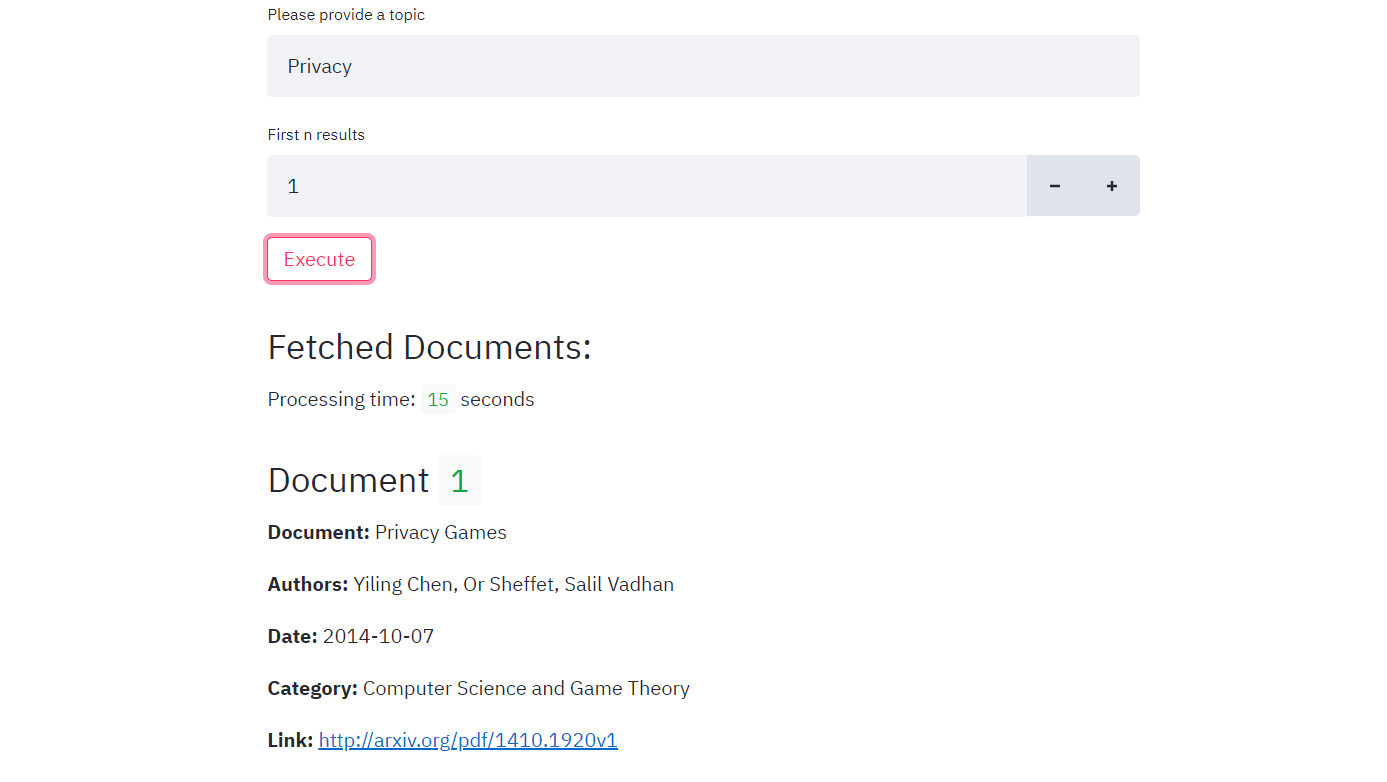}
\caption{Database Menu.}
\label{fig:db}
\end{figure}

For the topic of "Privacy", with a maximum of one article, the result is presented in Figure \ref{fig:db}.

Our solution for the cybersecurity use case performed admirably, by compiling a corpus of 821 articles on five of the hottest research topics in the field and by finding interesting answers to a set of significant questions regarding applications of AI to cybersecurity and the main challenges of current research. Regarding the extractive Q\&A pipeline, the RoBERTa model exhibited a notable adaptation capability since it was not retrained in the scope of the cybersecurity scientific domain.

\section{Conclusion}
\label{section:conclusion}

Given the amount of scientific articles that are published every year it is hard to find exactly what we are looking for when researching a particular topic. In this work, we have presented a software solution that aims to solve this problem. It comprises several advantageous features such as the continuous update of the search corpus by providing an easy-to-use integration with the arXiv.org API and the ability to find candidate answers extracted from the corpora of downloaded scientific publications by applying a combination of two NLP methods, TF-IDF and RoBERTa.

Furthermore, the introduced solution was showcased in the context of cybersecurity, a neoteric field of science with increasing interest. With a base corpus of 821 articles, the system was able to find proper answers to questions such as "What are the challenges of AI?", "What are the main challenges of cybersecurity research?" and "Which machine learning models are commonly used?" showing a great capability of generalization.

As future work, we will implement additional features regarding the document database management, expand the web crawler so that it can work with more scientific repositories and improve the document preprocessing step to make our search engine more efficient.

\bibliographystyle{ieeetr}
% splncs04
\bibliography{bibliography}

\end{document}